\documentclass{bmvc2k}

\usepackage{booktabs}
\usepackage{graphicx}
\usepackage{subfigure}
\usepackage{amsmath,amssymb}
\usepackage{color, colortbl}
\usepackage{wrapfig}

\title{Mean Box Pooling: A Rich Image
 Representation and Output Embedding for  the Visual Madlibs Task}

\addauthor{Ashkan Mokarian}{ashkan@mpi-inf.mpg.de}{1}
\addauthor{Mateusz Malinowski}{mmalinow@mpi-inf.mpg.de}{1}
\addauthor{Mario Fritz}{mfritz@mpi-inf.mpg.de}{1}

\addinstitution{
 Scalable Learning and Perception\\
 Max Planck Institute for Informatics\\
 Saarbrucken, Germany
}

\runninghead{Mokarian et. al.}{Mean Box Pooling}

\definecolor{LightCyan}{rgb}{0.88,1,1}

\newcommand{\bs}[1]{\ensuremath{\boldsymbol{#1}}}

\DeclareMathOperator* {\argmax}{arg\, max}

\begin{document}

\maketitle

\begin{abstract}
We present Mean Box Pooling, a novel visual representation that pools over CNN representations of a large number, highly overlapping object proposals.
We show that such representation together with nCCA, a successful multimodal embedding technique,  achieves state-of-the-art performance on the Visual Madlibs task. Moreover, inspired by the nCCA's objective function, we extend classical CNN+LSTM approach to train the network by directly maximizing the similarity between the internal representation of the deep learning architecture and candidate answers. Again, such approach achieves a significant improvement over the prior work that also uses CNN+LSTM approach on Visual Madlibs.
\end{abstract}

\section{Introduction}
\label{sec:intro}
\begin{wrapfigure}{r}{0.5\textwidth}
\begin{center}
\vspace{-35pt}
\includegraphics[width=0.48\textwidth]{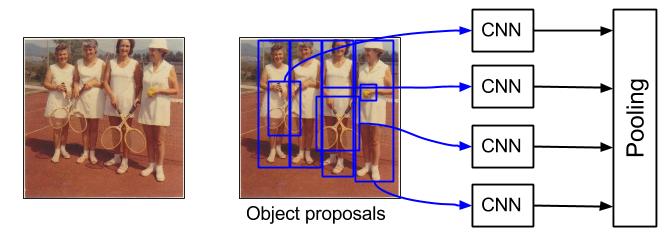}
\end{center}
\vspace{-14pt}
\caption{Illustration of proposed Mean Box Pooling representation.}
\vspace{-9pt}
\label{fig:proposal}
\end{wrapfigure}
Question answering about real-world
images is a relatively new research thread \cite{malinowski14nips,geman2015visual,malinowski14visualturing,antol2015vqa} that requires a chain of machine visual perception, natural language understanding, and deductive capabilities to successfully come up with an answer on a question about visual content. Although similar in nature to image description \cite{vinyals2014show,donahue2015long,karpathy15cvpr} it requires a more focused attention to details in the visual content, yet it is easier to evaluate different architectures on the task. Moreover, in contrast to many classical Computer Vision problems such as recognition or detection, the task does not evaluate any internal representation of methods, yet it requires a holistic understanding of the image. Arguably, it is also less prone to over-interpretations compared with the classical Turing Test \cite{turing1950computing, malinowski2015hard}.

To foster progress on this task, a few metrics and datasets have been proposed \cite{malinowski14nips, ren2015image, gao2015you, antol2015vqa}.
The recently introduced Visual Madlibs task \cite{visual_madlibs15} removes ambiguities in question or scene interpretations by introducing a multiple choice ``filling the blank'' task, where a machine has to complete the prompted sentence. Such completed sentence is next matched against four ground truth answers. Thanks to such a problem formulation, a traditional accuracy measure can be used to monitor the progress on this task. Due to its unambiguous evaluation, this work focuses on this task.

\paragraph{Contributions.}
We present two main contributions.\\
{\it Mean Box Pooling:} We argue for a rich image representation in the form of pooled representations of the objects. Although related ideas have been explored for visual question answering \cite{shih2015look}, and even have been used in Visual Madlibs \cite{visual_madlibs15}, we are first to show a significant improvement of such representation by using object proposals. More precisely, we argue for an approach that pools over a large number, highly overlapping  object proposals. This, arguably, increases the recall of extracting bounding boxes that describe an object, but also allows for multi-scale and multi-parts object representation. Our approach in the combination with the Normalized Correlation Analysis embedding technique improves on the state-of-the-art of the Visual Madlibs task.\\
{\it Text-Embedding Loss:} Motivated by the popularity of deep architectures for visual question answering, that combine a global CNN image representation with an LSTM \cite{hochreiter97nc} question representation \cite{malinowski2016ask,gao2015you,ren2015image,learning_to_answer_questions,yang2015stacked,xiong16dynamic,xu2015ask}, as well as the leading performance of nCCA on the multi-choice Visual Madlibs task \cite{visual_madlibs15}, we propose a novel extension of the CNN+LSTM architecture that chooses a prompt completion out of four candidates (see \autoref{fig:madlib_task}) by measuring similarities directly in the embedding space.  This contrasts with the prior approach of \cite{visual_madlibs15} that uses a post-hoc comparison between the discrete output of the CNN+LSTM method and all four candidates. To achieve this, we directly train an LSTM with a cosine similarity loss between the output embedding of the network and language representation of the ground truth completion. Such an approach integrates more tightly  with the multi-choice filling the blanks task, and significantly outperforms the prior CNN+LSTM method \cite{visual_madlibs15}.

\section{Related Work}
\label{sec:related_work}

Question answering about images is a relatively new task that switches focus from recognizing objects in the scene to a holistic ``image understanding''.
The very first work \cite{malinowski14nips} on this topic has considered real world indoor scenes with associated natural language questions and answers. Since then
different variants and larger datasets have been proposed: FM-IQA \cite{gao2015you}, COCO-QA \cite{ren2015image}, and VQA \cite{antol2015vqa}. Although answering questions on images is, arguably, more susceptible to automatic evaluation than the image description task \cite{vinyals2014show,karpathy15cvpr,donahue2015long}, ambiguities in the output space still remain. While such ambiguities can be handled using appropriate metrics \cite{vedantam2014cider, malinowski14nips, malinowski14visualturing, malinowski2016ask},  Visual Madlibs \cite{visual_madlibs15} has taken another direction, and handles them directly within the task. It asks machines to fill the blank prompted with a natural language description with a phrase chosen from four candidate completions (\autoref{fig:madlib_task}). In general, the phrase together with the prompted sentence should serve as the accurate description of the image.
With such problem formulation the
standard accuracy measure is sufficient to automatically evaluate the architectures.
The first proposed architecture \cite{malinowski14nips} to deal with the question answering about images task uses image analysis methods and a set of hand-defined schemas to create a database of visual facts. The mapping from questions to executable symbolic representations is done by a semantic parser \cite{liang2013learning}. Later deep learning approaches for question answering either generate  \cite{malinowski2016ask, gao2015you} answers or predict answers \cite{ren2015image,learning_to_answer_questions} over a fixed set of choices. Most recently, attention based architectures, which put weights on a fixed grid over the image, yield state of the art results \cite{yang2015stacked, xiong16dynamic,xu2015ask}. Another, more focused ``hard'' attention, has also been studied in the image-to-text retrieval scenario \cite{karpathy2014deep} as well as fine-grained categorization \cite{zhang2012pose}, person recognition \cite{oh15iccv} and zero-shot learning \cite{akata16cvpr}. Here representations are computed on objects, visual fragments or parts, that are further aggregated to form a visual representation. Closer to our work, \cite{shih2015look} use Edge Boxes \cite{zitnick2014edge} to form memories \cite{weston2014memory} consisting of different image fragments that are either pooled or ``softmax'' weighted in order to provide the final score. However, in contrast to \cite{shih2015look}, our experiments indicate a strong improvement by using object proposals.
While a majority of the most recent work on visual question answering combine LSTM \cite{hochreiter97nc} with CNN \cite{krizhevsky2012imagenet,szegedy2014going,simonyan2014very} by concatenation or summation or piece-wise multiplication, Canonical Correlation Analysis (CCA and nCCA) \cite{gong2014multi} have also been shown to be a very effective multimodal embedding technique \cite{visual_madlibs15}. Our work further investigates this embedding method as well as brings ideas from CCA over to an CNN+LSTM formulation.

\section{Method}
\label{sec:method}

\begin{figure}[t]
\begin{center}
\includegraphics[width=0.6\linewidth]{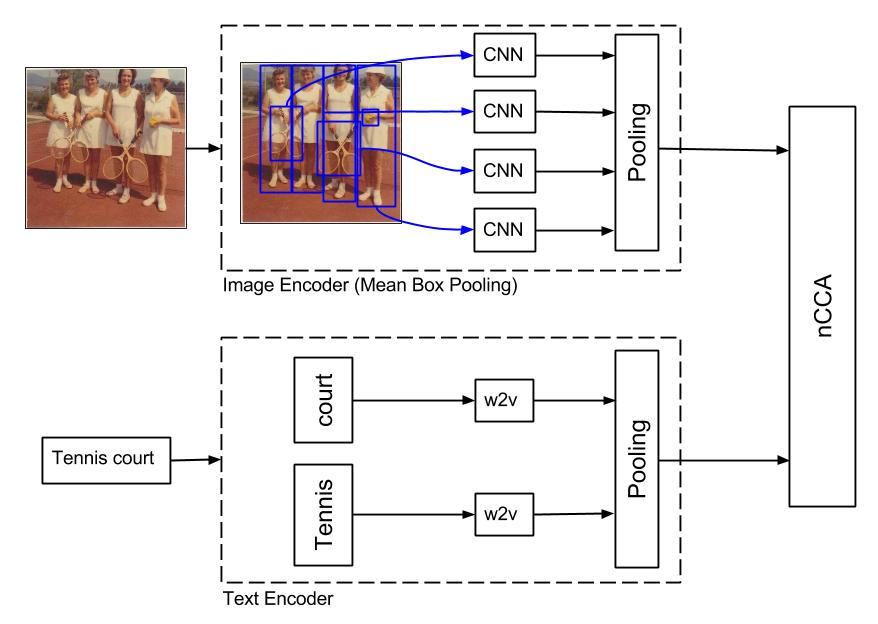}
\end{center}
\vspace{-15pt}
\caption{Overview of our full model, i.e. our proposed image representation using Mean Box Pooling, text encoding using average of Word2Vec representations, and normalized CCA for learning the joint space.}
\label{fig:whole_method}
\end{figure}

\begin{figure}[t]
\begin{center}
\includegraphics[width=0.7\linewidth]{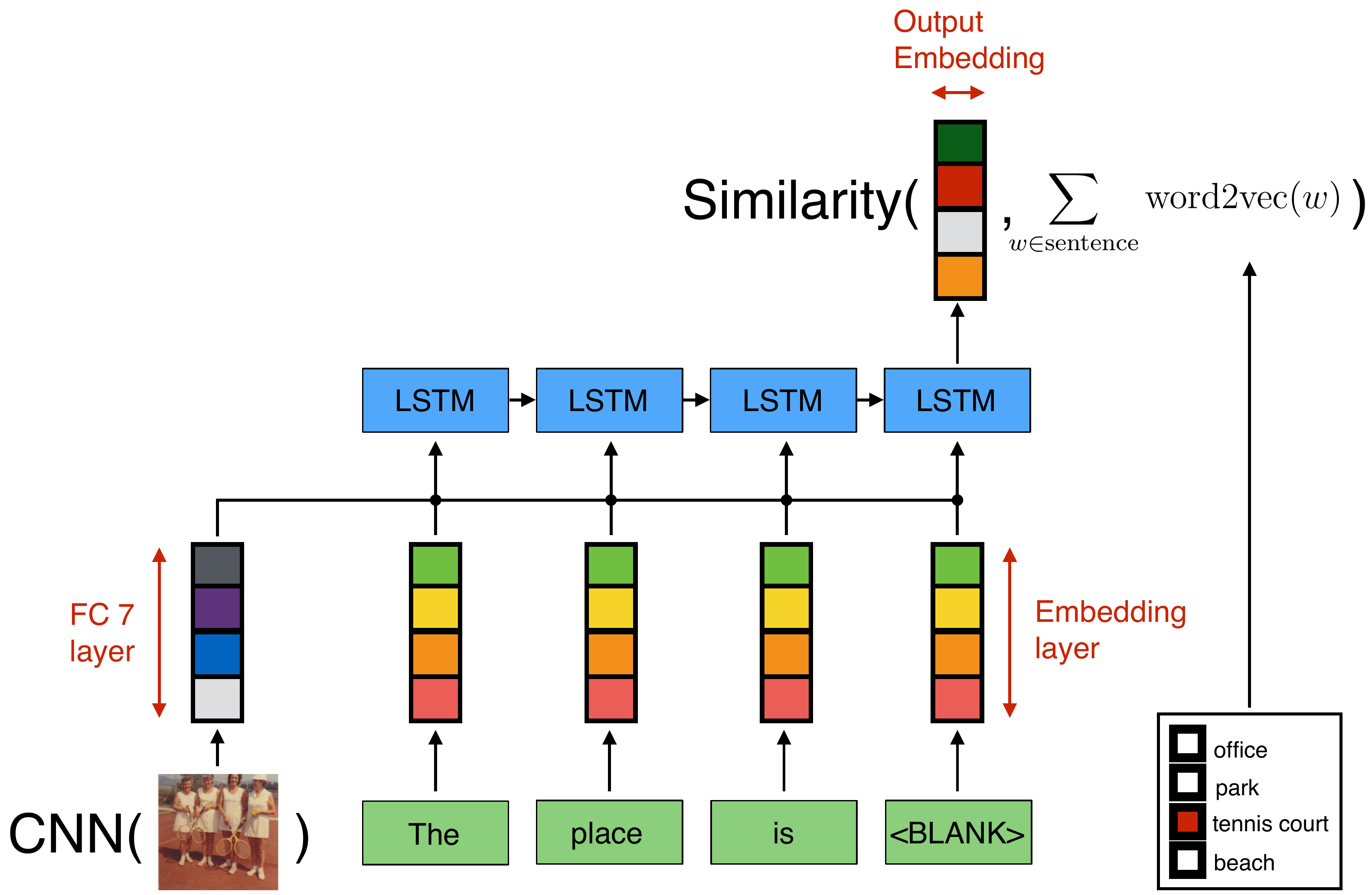}
\end{center}
\vspace{-15pt}
\caption{CNN+LSTM architecture that learns to choose the right answer directly in the embedding space. The output embedding is jointly trained with the whole architecture via backpropagation.}
\label{fig:lstm_embedding}
\end{figure}
 We use normalized CCA (nCCA) \cite{gong2014multi}
 to embed the textual embedding of answers and the visual representation of the image into a joint space, where candidate sentence completions are compared to the image. Furthermore, we also extend popular in the VQA community CNN+LSTM approach by learning to compare in the answer space.

In Section~\ref{sec:mbp}, we propose a richer representation  of the entire image obtained by pooling of CNN representations extracted from object proposals. Figure~\ref{fig:proposal} illustrates the proposed Mean Box Pooling image representation and Figure~\ref{fig:whole_method} illustrates our whole method. %
In Section~\ref{sec:me}, we describe nCCA approach to encode two modalities into a joint space in greater details.
In Section~\ref{sec:loss}, we also investigate a CNN+LSTM architecture.
Instead of generating a prompt completion that is next compared against candidate completions in a post-hoc process, we propose to choose a candidate completion by directly comparing candidates in the embedding space. This puts CNN+LSTM approach closer to nCCA with a tighter integration with the multi-choice Visual Madlibs task. This approach is depicted in Figure~\ref{fig:lstm_embedding}.
\subsection{Mean Box Pooling Image Representation}\label{sec:mbp}
Figure~\ref{fig:proposal} illustrates our proposed image representation, which starts from extracting object proposals from the raw image. Next such object proposals are encoded via a CNN, and pooled in order to create a feature vector representation of the image.

\paragraph{Extracting Region Proposals.}
Since questions are mainly about salient parts of the image, it seems reasonable to use object detection in order to extract such parts from the image. At the same time, however, it is important to not miss any object in the image. Moreover, arguably, sampling a context of the objects and capturing multi-scale, multi-parts properties seem to be important as well.
 Given all these reasons, we choose to use Edge Boxes \cite{zitnick2014edge} in order to generate a set of object bounding box proposals for feature extraction.

Edge Boxes extract a number of bounding boxes along with a score for each bounding box that is interpreted as a confidence score that the bounding box contains an object. In our study, two hyper parameters are important: Non-Maxima Suppression and the number of proposals.
The latter defines how many object proposals we want to maintain and hence implicitly influence recall of the proposals, while the former defines a threshold $\beta$ such that all predicted bounding boxes with the intersection over union greater than $\beta$ are removed. In practice, the lower the $\beta$ the more spread the object proposals are.

\paragraph{Feature Extraction.}
Once the object proposals are extracted, we use output of the ``fc7'' layer of the VGG network \cite{simonyan2014very} on the extracted image crops to encode the proposals. VGG is among the best performing recognition architectures on the large scale object recognition task \cite{ILSVRCarxiv14}.

\paragraph{Pooling for Image Representation.}
Our final image representation is constructed by pooling the encoded object proposals together with the global image representation. Since we do not want to associate any particular order over the extracted object proposals,  we investigate popular order-less pooling schemes.

\subsection{Pooling for Answer Representation.}
\label{sec:language_representation}
We encode each word in the answer with a $300$ dimensional word embedding \cite{mikolov2013distributed}. The embedded words are next mean pooled to form a vector representation of the answer. Note that, we do not encode prompts as they follow the same pattern for each Visual Madlibs category.

\subsection{Multimodal Embedding}
\label{sec:me}
We use the Normalized Canonical Correlation Analysis (nCCA)  to learn a mapping from two modalities: image and textual answers, into a joint embedding space. This embedding method has shown outstanding performance on the Visual Madlibs task \cite{visual_madlibs15}. At the test time, given the encoded image, we choose an answer (encoded by the mean pooling over word2vec words representations) from the set of four candidate answers that is the most similar to the encoded image in the multimodal embedding space.
Formally, the Canonical Correlation Analysis (CCA) maximizes the cosine similarity between two modalities (also called views) in the embedding space, that is:
\begin{align*}
  W_1^{\ast}, W_2^{\ast} =  \argmax_{\bs{W}_1, \bs{W}_2} \;\;\; & \text{tr}(\hat{X}^T \hat{Y}) \\
  \text{subject to\;\;\;} & \hat{X}^T \hat{X} = \hat{Y}^T \hat{Y} = I
\end{align*}
where $tr$ is the matrix trace,  $\hat{X} := X W_1$, $\hat{Y} := Y W_2$, and $X,Y$ are two views (encoded images, and textual answers in our case).
Normalized Canonical Correlation Analysis (nCCA) \cite{gong2014multi} has been reported to work significantly better than the plain CCA. Here, columns of the projection matrices $W_1$ and $W_2$ are scaled by the p-th power (p=4) of the corresponding eigen values.
The improvement is consistent with the findings of \cite{visual_madlibs15}, where nCCA performs better than CCA by about five percentage points in average on the hard task.

\subsection{CNN+LSTM with Text-Embedding Loss}\label{sec:loss}
We present our novel architecture that extends prior approaches on question answering about images \cite{malinowski2016ask,gao2015you,ren2015image,learning_to_answer_questions,yang2015stacked,xiong16dynamic,xu2015ask} by learning similarity between candidate labels and internal output embedding of the neural network. \autoref{fig:lstm_embedding} depicts our architecture. Similarly to prior work, we encode an image with a CNN encoder that is next concatenated with (learnable) word embeddings of the prompt sentence, and fed to a recurrent neural network. We use a special `<BLANK>' token to denote the empty blank space in the image description. On the other side, for each completion candidate $s$ we compute its representation by averaging over word2vec \cite{mikolov2013distributed} representations of the words contributing to $s$. However, in contrast to the prior work \cite{visual_madlibs15}, instead of comparing the discrete output of the network with the representation of $s$, we directly optimize an objective in the embedding space. During training we maximize the similarity measure between the output embedding and the representation of $\sigma$ by optimizing the following objective:
\begin{align*}
  \Theta^{\ast} = \argmax_{\Theta} \sum_{i} \frac{\text{embedding}(\bs{x}_i; \Theta)^T \left(\sum_{w \in \hat{s}_i} \text{word2vec}(w)\right)}{||\text{embedding}(\bs{x}_i; \Theta)|| \; ||\sum_{w \in \hat{s}_i} \text{word2vec}(w)||},
\end{align*}
which is a cosine similarity between the representation of the available during the training correct completion $\hat{s}_i$, and an output embedding vector of the i-th image-prompt training instance $\bs{x}_i$; $\Theta$ denotes all the parameters of the architecture. At test time,  we choose a completion $\hat s$ by:
\begin{align*}
  \hat s = \argmax_{s \in \mathcal{S}} \frac{\text{embedding}(\bs{x}; \Theta^{\ast})^T \left(\sum_{w \in s}\text{word2vec}(w)\right)}{|| \text{embedding}(\bs{x}; \Theta^{\ast}) || \; || \sum_{w \in s} \text{word2vec}(w) ||},
\end{align*}
where $\mathcal{S}$ denotes a set of available candidate prompt completions,  $\bs{x}$ is the image-prompt pair fed to the network, and  $\Theta^{\ast}$ denotes all the learnt parameters.

\section{Experimental Results}
\label{sec:results}
\begin{figure}
\center
\includegraphics[width=0.8\textwidth]{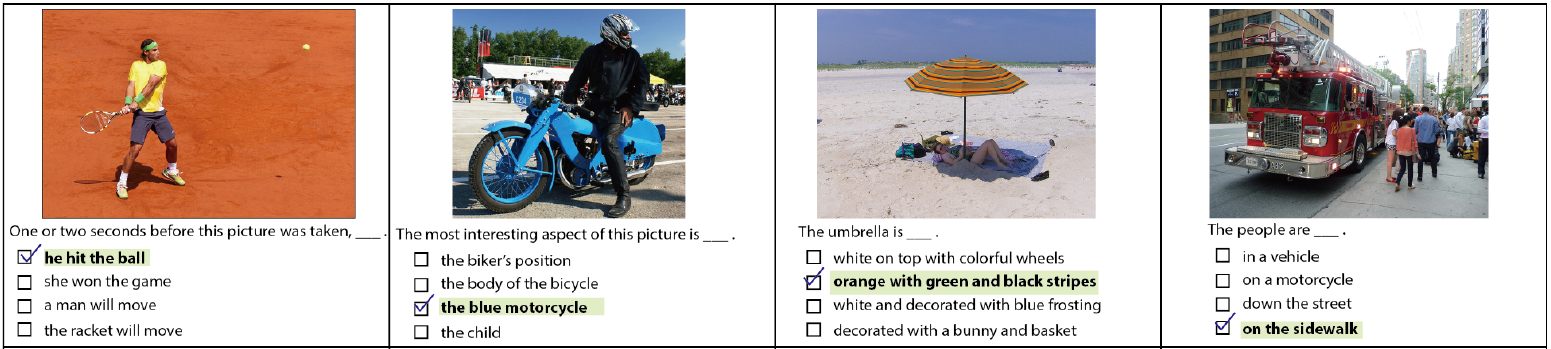}
\vspace{-5pt}
\caption{Some examples of the multi-choice filling the blank Visual Madlib task \cite{visual_madlibs15}.}
\label{fig:madlib_task}
\end{figure}

We evaluate our method on the multiple choice task of the Visual Madlibs dataset. The dataset consists of about 360k descriptions, spanning 12 different categories specified by different types of templates, of about 10k images.
The selected images from the MS COCO dataset comes with  rich annotations.
In the multi-choice scenario a textual prompt is given (every category follows the same, fixed template) with a blank to be filled, together with 4 candidate completions (see \autoref{fig:madlib_task}). Every category represents a different type of question including scenes, affordances, emotions, or activities (the full list is shown in the first column of \autoref{tab:meanEB:nCCA-nms}). Since each category has fixed prompt, there is no need to include the prompt in the modeling given the training is done per each category.
Finally, Visual Madlibs considers an easy and difficult tasks that differ in how the negative 3 candidate completions (distractors) are chosen. In the easy task, the distractors are randomly chosen from three descriptions of the same question type from other images. In the hard task, 3 distractors are chosen only from these images that contain the same objects as the given question image, and hence it requires a more careful and detailed image understanding. We use ADAM gradient descent method \cite{kingma2014adam} with default hyper-parameters.
\paragraph{Different Non Maxima Suppression Thresholds.}
\autoref{tab:meanEB:nCCA-nms} shows the accuracy of our CCA model with mean edge-box pooling for two different Non Maxima Suppression (NMS) thresholds $\beta$. In the experiments we use up to $100$ object proposals, as we have observed saturation for higher numbers.  From \autoref{tab:meanEB:nCCA-nms}, we can see that for both tasks, easy and hard, higher NMS thresholds are preferred. More precisely, the threshold $0.75$ outperforms $0.30$ in average by  $2.4$ percentage points for the easy task, and by $1.8$ percentage points for the hard task.
We have also experimented with max pooling, but  mean pooling has performed by about $0.5$ percentage points better in average in all our experiments. The experiments, counterintuitively, suggest that many selected bounding boxes with high overlap are still beneficial in achieving better performance. Further experiments use mean pooling and the NMS threshold $0.75$.

\tabcolsep 3pt
\begin{table}[t]
  \centering
  \scalebox{0.8}{
  \begin{tabular}{ l  c c | c c}
  \multicolumn{3}{r}{\textbf{Easy Task}} & \multicolumn{2}{|c}{\textbf{Hard Task}} \\
    \toprule
    NMS thresholds: & 0.30 & 0.75 & 0.30 &0.75  \\
    \midrule
    Image's scenes     & 84.6 &\textbf{86.2} & 67.8 &\textbf{69.0} \\
    Image's emotion    & 52.0   &\textbf{52.5} & 38.8 &\textbf{39.4} \\
    Image's past       & 78.9 &\textbf{80.8} & 53.0  &\textbf{54.6} \\
    Image's future     & 78.6 &\textbf{81.1} & 55.2 &\textbf{56.1} \\
    Image's interesting& 76.8 &\textbf{78.2} & 53.5 &\textbf{54.2}  \\
    Object's attribute & 60.4 &\textbf{62.4} & 43.1 &\textbf{45.7}  \\
    Object's affordance  & 80.4 &\textbf{83.3} & 62.5 &\textbf{63.6} \\
    Object's position  & 73.5 &\textbf{77.4} & 53.6 &\textbf{56.3} \\
    Person's attribute & 53.1 &\textbf{56.0} & 42.0   &\textbf{44.2}    \\
    Person's activity  & 79.9 &\textbf{83.0} & 63.2 &\textbf{65.5}   \\
    Person's location  & 82.4 &\textbf{84.3} & 63.8 &\textbf{65.2} \\
    Pair's relationship  & 71.0   &\textbf{75.3}& 51.8 &\textbf{55.7} \\
    \rowcolor{LightCyan}
    Average    & 72.6 &\textbf{75.0}& 54.0   &\textbf{55.8}  \\
    \bottomrule
  \end{tabular}
  }
    \vspace{0.1em}
  \caption[Comparison of different NMS values for EdgeBoxes and different pooling methods]{Accuracies computed for different Non Maxima Suppression thresholds (NMS) on the easy and hard tasks of the Visual Madlibs dataset. Mean pooling and 100 object proposals are used in the experiments. Results in $\%$.}
    \vspace{-1.0em}
  \label{tab:meanEB:nCCA-nms}
\end{table}

\paragraph{Different number of object proposals.}
The maximal number of object proposals is the second factor of Edge Boxes that we study in this work. A larger number of proposals tend to cover a larger fraction of the input image. Moreover, the higher number together with the higher NMS threshold can assign proposals to both an object, and its parts, effectively forming a multi-scale and multi-parts object representation.
\autoref{tab:meanEB:nCCA-numEB} shows the accuracy  of our model with different number of Edge Box proposals. The experiments suggest using  a larger numbers of proposals, however the gain diminishes with the larger numbers.

\tabcolsep 3pt
\begin{table}[t]
  \centering
  \scalebox{0.8}{
  \begin{tabular}{ l  c c c c}
  \multicolumn{5}{c}{\textbf{Easy Task}} \\
    \toprule
    Number of proposals & 10 & 25 & 50 & 100 \\
    \midrule
    Scenes     & 84.5 & 85.5 & 86.0 & \textbf{86.2}\\
    Emotion    & 49.9 & 51.6 & 52.1 & \textbf{52.5}\\
    Past       & 78.7 & 80.0 & 80.6 & \textbf{80.8}\\
    Future     & 78.7 & 79.7 & 80.7 & \textbf{81.1}\\
    Interesting& 75.4 & 77.2 & 77.9 & \textbf{78.2}\\
    Obj. attr. & 59.0  & 60.9 & 61.7 & \textbf{62.4}\\
    Obj. aff.  & 81.2 & 82.4 & 83.0 & \textbf{83.3}\\
    Obj. pos.  & 75.4 & 76.6 & 77.4 & \textbf{77.5}\\
    Per. attr. & 51.4 & 53.3 & 55.0 & \textbf{56.0}\\
    Per. act.  & 80.7 & 82.2 & 82.9 & \textbf{83.0}\\
    Per. loc.  & 82.9 & 83.9 & 84.0 & \textbf{84.3}\\
    Pair's rel.  & 72.4 & 73.9 & 74.6 & \textbf{75.3}\\
    \rowcolor{LightCyan}
    Average       & 72.5 & 73.9 & 74.7 & \textbf{75.0}\\
    \bottomrule
  \end{tabular}
  }
  \quad
  \scalebox{0.8}{
  \begin{tabular}{l c c c c}
    \multicolumn{5}{c}{\textbf{Hard Task}} \\
    \toprule
    Number of proposals & 10 & 25 & 50 & 100 \\
    \midrule
    Scenes     & 68.0   & 68.6 & 68.9 & \textbf{69.0}\\
    Emotion    & 37.9 & 38.1 & 38.8 & \textbf{39.4}\\
    Past       & 52.8 & 53.9 & 54.3 & \textbf{54.6}\\
    Future     & 54.4 & 55.0 & 55.8 & \textbf{56.1}\\
    Interesting& 51.9 & 53.6 & 53.7 & \textbf{54.2}\\
    Obj. attr. & 43.7 & 44.0 & 44.9 & \textbf{45.7}\\
    Obj. aff.  & 62.4 & 63.0 & 63.4 & \textbf{63.6}\\
    Obj. pos.  & 55.1 & 55.5 & 56.3 & \textbf{56.3}\\
    Per. attr. & 41.6 & 42.2 & 43.0 & \textbf{44.2}\\
    Per. act.  & 63.7 & 64.7 & 65.3 & \textbf{65.5}\\
    Per. loc.  & 64.2 & 64.8 & 64.8 & \textbf{65.2}\\
    Pair's rel.  & 53.6 & 54.5 & 54.9 & \textbf{55.7}\\
    \rowcolor{LightCyan}
    Average       & 54.1 & 54.8 & 55.3 & \textbf{55.8}\\
    \bottomrule
  \end{tabular}
  }
  \caption[Accuracies computed for different number of Edge Box proposals]{Accuracies computed for different number of Edge Box proposals on the easy and hard tasks of the Visual Madlibs dataset. The NMS threshold 0.75 and mean-pooling is used for all the experiments. Results in $\%$.}
  \label{tab:meanEB:nCCA-numEB}
\end{table}

\paragraph{Comparison to the state-of-the-art.}
Guided by the results of the previous experiments, we compare nCCA that uses Edge Boxes object proposals (nCCA (ours)) with the state-of-the-arts on Visual Madlibs (nCCA \cite{visual_madlibs15}). Both models use the same VGG Convolutional Neural Network \cite{simonyan2014very} to encode images (or theirs crops), and word2vec to encode words. The models are trained per category (a model trained over all the categories performs inferior on the hard task \cite{visual_madlibs15}). As \autoref{tab:meanEB:nCCA-stateofart} shows using a large number of object proposals improves over global, full frame nCCA by $5.9$ percentage points on the easy task, and about $1.4$ percentage points on the difficult task in average. However, our nCCA also consistently outperforms state-of-the-art on every category except the `Scenes' category. This suggests that better localized object oriented representation is beneficial.
However, Edge Boxes only roughly localize objects.
This naturally leads to the following question if better localization helps. To see the limits, we compare nCCA (ours) against nCCA (bbox) \cite{visual_madlibs15} that crops over ground truth bounding boxes from MS COCO segmentations and next averages over theirs representations (Table 3 in \cite{visual_madlibs15} shows that ground truth bounding boxes outperforms automatically detected bounding boxes, and hence they can be seen as an upper bound for a detection method trained to detect objects on MS COCO). Surprisingly, nCCA (ours) outperforms nCCA (bbox) by a large margin as \autoref{tab:meanEB:nCCA-stateofart-bbox} shows. Arguably, object proposals have better recall and captures multi-scale, multi-parts phenomena.

\tabcolsep 3pt
\begin{table}[t]
  \centering
  \scalebox{0.8}{
  \begin{tabular}{ l  c c | c c}
  \multicolumn{3}{c}{\textbf{Easy Task}} & \multicolumn{2}{|c}{\textbf{Hard Task}} \\
    \toprule
    & nCCA (ours) & nCCA \cite{visual_madlibs15} & nCCA (ours) & nCCA \cite{visual_madlibs15}   \\
    \midrule
    Scenes     & 86.2 & \textbf{86.8} & 69.0 & \textbf{70.1} \\
    Emotion    & \textbf{52.5} & 49.2 & \textbf{39.4} & 37.2 \\
    Past       & \textbf{80.8} & 77.5 & \textbf{54.6} & 52.8 \\
    Future     & \textbf{81.1} & 78.0 & \textbf{56.1} & 54.3 \\
    Interesting& \textbf{78.2} & 76.5 & \textbf{54.2} & 53.7 \\
    Obj. attr. & \textbf{62.4} & 47.5 & \textbf{45.7} & 43.6\\
    Obj. aff.  & \textbf{83.3} & 73.0 & \textbf{63.6} & 63.5 \\
    Obj. pos.  & \textbf{77.5} & 65.9 & \textbf{56.3} & 55.7\\
    Per. attr. & \textbf{56.0} & 48.0 & \textbf{44.2} & 38.6 \\
    Per. act.  & \textbf{83.0} & 80.7 & \textbf{65.5} & 65.4 \\
    Per. loc.  & \textbf{84.3} & 82.7 & \textbf{65.2} & 63.3 \\
    Pair's rel.  & \textbf{75.3} & 63.0 & \textbf{55.7} & 54.3\\
    \rowcolor{LightCyan}
    Average   & \textbf{75.0} & 69.1 & \textbf{55.8} & 54.4 \\
    \bottomrule
  \end{tabular}
  }
    \vspace{0.1em}
  \caption[Comparison of our proposed visual representation to others on Visual Madlibs]{Accuracies computed for different approaches on the easy and hard tasks. nCCA (ours) uses the representation with object proposals (NMS 0.75, and 100 proposals with mean-pooling). nCCA uses the whole image representation. Results in $\%$.}
    \vspace{-1.0em}
  \label{tab:meanEB:nCCA-stateofart}
\end{table}

\tabcolsep 3pt
\begin{table}[t]
  \centering
  \scalebox{0.8}{
  \begin{tabular}{ l  c c | c c}
  \multicolumn{3}{c}{\textbf{Easy Task}} & \multicolumn{2}{|c}{\textbf{Hard Task}} \\
    \toprule
    & nCCA (ours) &  nCCA (bbox) \cite{visual_madlibs15} & nCCA (ours) &  nCCA (bbox) \cite{visual_madlibs15} \\
    \midrule
    Obj. attr. & \textbf{62.4} & 54.7 & 45.7 &  \textbf{49.8} \\
    Obj. aff.  & \textbf{83.3} & 72.2 & \textbf{63.6} &  63.0 \\
    Obj. pos.  & \textbf{77.5} & 58.9 & \textbf{56.3} &  50.7\\
    Per. attr. & \textbf{56.0} & 53.1 & 44.2 &  \textbf{46.1} \\
    Per. act.  & \textbf{83.0} & 75.6 & \textbf{65.5} &  65.1 \\
    Per. loc.  & \textbf{84.3} & 73.8 & \textbf{65.2} & 57.8 \\
    Pair's rel.  & \textbf{75.3} & 64.2 & 55.7 & \textbf{56.5} \\
    \rowcolor{LightCyan}
    Average   & \textbf{74.5}  & 64.6 & \textbf{56.6} & 55.6\\
    \bottomrule
  \end{tabular}
  }
    \vspace{0.1em}
  \caption[Comparison of our proposed visual representation to others on Visual Madlibs]{Accuracies computed for different approaches on the easy and hard task. nCCA (ours) uses the representation with object proposals (NMS 0.75, and 100 proposals with mean-pooling). nCCA(bbox)  mean-pools over the representations computed on the available ground-truth bounding boxes both at train and test time. The averages are computed only over 7 categories. Results in $\%$.}
    \vspace{-1.0em}
  \label{tab:meanEB:nCCA-stateofart-bbox}
\end{table}

\paragraph{CNN+LSTM with comparison in the output embedding space.}
On one hand nCCA tops the leaderboard on the Visual Madlibs task \cite{visual_madlibs15}. On the other hand, the largest body of work on the  question answering about images  \cite{malinowski14nips,gao2015you,ren2015image,antol2015vqa} combines a CNN with an LSTM \cite{malinowski2016ask,gao2015you,ren2015image,learning_to_answer_questions,yang2015stacked,xiong16dynamic,xu2015ask}. We hypothesize that, likewise to nCCA, in order to choose a completion of the prompt sentence out of four candidates, the comparison between the candidate completions should be directly done in the output embedding space. This contrasts to a post-hoc process used in \cite{visual_madlibs15} where an image description architecture (CNN+LSTM) first generates a completion that is next compared against the candidates in the word2vec space (see \autoref{sec:method} for more details). Moreover, since the ``Ask Your Neurons'' architecture \cite{malinowski2016ask} is more suitable for the question answering task, we extend that method to do comparisons directly in the embedding space (``Embedded CNN+LSTM'' in \autoref{tab:LSTM_CNN}). Note that, here we feed the sentence prompt to LSTM even though it is fixed per category. \autoref{tab:LSTM_CNN} shows the performance of different methods. Our ``Embedded CNN+LSTM'' outperforms other methods on both tasks confirming our hypothesis. ``Ask Your Neurons'' \cite{malinowski2016ask} is also slightly better than the original CNN+LSTM \cite{visual_madlibs15} (on the 10 categories that the results for CNN+LSTM are available it achieves $49.8\%$ accuracy on the easy task, which is $2.1$ percentage points higher than CNN+LSTM).

\tabcolsep 3pt
\begin{table}[t]
  \centering
  \scalebox{0.7}{
  \begin{tabular}{ l  c c c | c c}
  \multicolumn{4}{c}{\textbf{Easy Task}} & \multicolumn{2}{|c}{\textbf{Hard Task}} \\
    \toprule
    & Embedded        & Ask Your                         & CNN+LSTM  & Embedded  &  CNN+LSTM  \\
    & CNN+LSTM (ours) & Neurons \cite{malinowski2016ask} & \cite{visual_madlibs15} & CNN+LSTM (ours) & \cite{visual_madlibs15} \\
    \midrule
    Scenes     & \textbf{74.7} & 70.6 & 71.1 & \textbf{62.1} & 60.5 \\
    Emotion    & \textbf{36.2} & 35.7 & 34.0 & \textbf{34.3} & 32.7 \\
    Past       & \textbf{46.8} & 44.9 & 35.8 & \textbf{42.5} & 32.0\\
    Future     & \textbf{48.1} & 41.2 & 40.0 & \textbf{41.4} & 34.3\\
    Interesting& \textbf{49.9} & 49.1 & 39.8 & \textbf{40.1} & 33.3\\
    Obj. attr. & \textbf{46.5} & 45.5 & 45.4 & \textbf{40.6} & 40.3\\
    Obj. aff.  & \textbf{68.5} & 64.3 & -    & \textbf{86.4} & - \\
    Obj. pos.  & \textbf{53.3} & 49.5 & 50.9 & \textbf{45.0} & 44.9\\
    Per. attr. & \textbf{40.7} & 39.9 & 37.3 & \textbf{40.0} & 35.1\\
    Per. act.  & \textbf{64.1} & 62.6 & 63.7 & \textbf{53.7} & 53.6\\
    Per. loc.  & \textbf{61.5} & 59.1 & 59.2 & \textbf{51.4} & 49.3\\
    Pair's rel.& \textbf{66.2} & 60.1 & -    & \textbf{54.5} & -\\
    \rowcolor{LightCyan}
    Average   & \textbf{54.7} & 51.9 & 47.7 & \textbf{49.3} & 41.7 \\
    \bottomrule
  \end{tabular}
  }
    \vspace{0.1em}
  \caption[Comparison of our proposed visual representation to others on Visual Madlibs]{Comparison between our Embedded CNN+LSTM approach that  computes the similarity between input and candidate answers in the embedding space, and the plain CNN+LSTM original approach from \cite{visual_madlibs15}. Since the accuracies of CNN+LSTM \cite{visual_madlibs15} are unavailable for two categories, we report average over 10 categories in this case. Results in $\%$.}
    \vspace{-1.0em}
  \label{tab:LSTM_CNN}
\end{table}

\section{Conclusion}
\label{sec:conclusion}
We study an image representation formed by averaging over representations of object proposals, and show its effectiveness through experimental evaluation on the Visual Madlibs dataset \cite{visual_madlibs15}. We achieve state of the art performance on the multi-choice ``filling the blank'' task. We have also shown and discussed effects of different parameters that affect how the proposals are obtained.
Surprisingly, the larger number of proposals the better overall performance. Moreover, the model benefits even from highly overlapping proposals. Such model even outperforms the prior work that uses ground truth bounding boxes from the MS COCO dataset. The proposed representation can be considered as a valid alternative to `soft' attention representations such as implemented in recent work of visual question answering using memory networks \cite{yang2015stacked}.
Due to its popularity on question answering about images tasks, we also investigate a CNN+LSTM approach that chooses a prompt completion candidate by doing comparisons directly in the embedding space. This approach contrasts with a post-hoc solution of the previous work allowing for a tighter integration of the model with the multi-choice task.

\paragraph{Acknowledgements:} This work is supported by the German Research Foundation (DFG) under the SFB/CRC 1223.

\bibliography{mateusz}
\end{document}